\documentclass[runningheads]{llncs}
\usepackage{graphicx}
\usepackage{hyperref}
\usepackage{xcolor}
\usepackage{placeins}
\newcommand{\cm}[1]{\textcolor{black}{#1}}

\newcommand{\nt}[1]{\textcolor{black}{#1}}
\newcommand{\add}[1]{\textcolor{black}{#1}} 

\usepackage{amsmath}
\usepackage{latexsym}
\usepackage{multirow}
\usepackage{lipsum}
\usepackage{graphicx}
\usepackage{amssymb}
\usepackage{cite}
\usepackage{multirow}
\usepackage{booktabs}
\usepackage{tabularx}
\usepackage{paralist}

\begin{document}
\title{\cm{Near-}Zero-Shot Suggestion Mining  \\ with a Little Help from WordNet}
%


\author{Anton Alekseev\inst{1,2}\orcidID{0000-0001-6456-3329}, Elena Tutubalina\inst{3,4,5}\orcidID{0000-0001-7936-0284}, Sejeong Kwon\inst{6} and Sergey Nikolenko\inst{1}\orcidID{0000-0001-7787-2251}}
\institute{
Steklov Mathematical Institute at St.~Petersburg, Russia \and 
St. Petersburg University, Russia \and
Kazan (Volga Region) Federal University, Kazan, Russia \and 
HSE University, Russia \and 
Sber AI, Russia \and 
Samsung Research, Korea}

%
%
%
\maketitle              

\begin{abstract}

In this work we explore the \emph{constructive} side of online reviews: advice, tips, requests, and suggestions that users provide about goods, venues, services and other items of interest. 
To reduce training costs and annotation efforts needed to build a 
classifier for a specific label set, we present and evaluate
several entailment-based zero-shot approaches to suggestion classification in a \textit{label-fully-unseen} fashion.
In particular, we introduce the strategy of assigning target class labels to sentences \add{in English language} 
with user intentions, which significantly improves prediction quality.
\nt{The proposed strategies are evaluated with a comprehensive experimental study that validated our results both quantitatively and qualitatively.}




\end{abstract}

\section{Introduction}\label{sec:introduction}

Online user reviews often provide feedback that extends much further than just the overall or aspect-specific sentiment. 
Users can \nt{describe their experience in detail and, in particular,}
provide advice tips or suggestions that can 
\nt{be useful both to other users and service providers.
One of the most valuable entities both for peer users and the reviewed object's owners/manufacturers/developers/sellers/etc. are 
reviews with user-generated suggestions that}
help other users make informed decisions and choices, while providers responsible for reviewed items get more specific advice on which modifications to make or on the selling strategy.

\nt{Therefore,}
with growing volumes of opinions and reviews posted online, it is important to develop an effective way to extract advice/tips/suggestions for highlighting or aggregation.
The task of automatic identification
of suggestions in a given text is known as
\emph{suggestion mining}~\cite{brun2013suggestion}.
It is usually defined as a sentence classification task: 
each sentence of a review is assigned a class of either ``suggestion'' or ``non-suggestion''~\cite{negi2016study}. 
\add{E.g. sentences from the dataset~\cite{negi2019semeval} ``Having read through the cumbersome process, I am more incline to build FREE apps with subscription rather than getting any money through the Marketplace'' and ``Why can't Microsoft simplify it?'' do not propose any specific changes/improvements to the systems the users are writing about, while  ``Even something as simple as ctrl+S would be a godsend for me'' does. Hence, the latter can be called a \emph{suggestion} and the first two can not. Also note that this suggestion example is not formulated as a direct request, which demonstrates that the task is harder than simply writing a set of lexical patterns.}

Suggestion mining finds multiple applications in a number of industries, from consumer electronics to realty. In each of these domains, however, suggestions and tips are proposed in different ways, so NLP approaches to processing online reviews differ across domains as well.
That is why training domain-independent embeddings, e.g., for the task of cross-domain suggestion mining is a hard challenge; see, e.g., \textit{SemEval-2019 Task 9B}~\cite{negi2019semeval}. 
Since most state-of-the-art models use machine learning to tune their parameters, their performance---and relevance to real-world implementations---is highly dependent on the dataset on which they are trained~\cite{wu2015collaborative,chen2018multinomial}.

In this work, we attempt to avoid this problem by focusing on the \emph{\cm{(near-)} zero-shot learning} approach, which does not require any training data.
We extend the methodology recently proposed by~\cite{yin2019benchmarking} to this new task. Our approach is based on natural language inference 
\nt{that includes a sentence to be classified as premise and a sentence describing a class as hypothesis.}

\nt{In order to make the proposed methods more universally applicable across different domains, and following the methodology proposed in~\cite{yin2019benchmarking},} 
we explore the extent to which suggestion classification can be solved with zero-shot approaches. 
For that purpose, we use the BART model~\cite{lewis2019bart} pretrained for natural language inference~\cite{mnli}. 
The main contributions of this work include:
\begin{enumerate}[(1)]
\item new WordNet-based modifications to the basic approach to zero-shot \textit{label-fully-unseen} classification of suggestions \nt{(\textit{near}-zero-shot classification)} and
\item established benchmarks in zero-shot learning for suggestion classification in English.
\end{enumerate}
We also test whether the best method is easily transferable to other domains and report negative results.

\nt{The paper is structured as follows. Section~\ref{sec:relatedwork} surveys related work in opinion mining and natural language inference. In Section~\ref{sec:data} we describe the datasets used 
for this work. Section~\ref{sec:methods} introduces the proposed approaches to zero-shot and near-zero-shot learning, Section~\ref{sec:results} presents and discusses the numerical results of our experiments, and Section~\ref{sec:conclusion} concludes the paper.}
\section{Related Work}\label{sec:relatedwork}

The two main topics related to this work are
\begin{inparaenum}[(1)]
\item opinion mining, \nt{especially on the side of datasets collected and used in this work}, 
and \item natural language inference,
\nt{where most of the methods applied here come from}. 
\end{inparaenum} 
\nt{In this section, we briefly survey state of~the art in both topics.}

In opinion mining, existing research has been focused on expressions of opinion that convey positive or negative sentiments \cite{pang2008opinion,liu2010sentiment,feldman2013techniques,wu2015collaborative,chaturvedi2018distinguishing,tutubalina2015inferring,liu2017many}. For example, two widely studied topics \cm{in} 
text classification are whether a given text is
\begin{enumerate}[(i)]
\item positive or negative, \nt{a problem known as \emph{sentiment analysis}}, and 
\item subjective or objective, which distinguishes subjective and fact-based opinions; see comprehensive surveys of both topics in~\cite{pang2008opinion,tsytsarau2012survey,liu2010sentiment,chaturvedi2018distinguishing}.
\end{enumerate}

A fact-based opinion is a regular or comparative opinion implied in an objective or factual statement~\cite{liu2017many}.  In a pioneering work, Pang and Lee~\cite{pang2004sentimental} presented a preprocessing filter that labels text sentences as either subjective or objective. It removes the latter and uses subjective sentences for sentiment classification of movie reviews. Raychev and Nakov~\cite{raychev2009language} proposed a novel approach that assigns position-dependent weights to words or word bigrams based on their likelihood of being subjective; their evaluation shows a na\"ive Bayes classifier on a standard dataset of movie reviews on par with state-of-the-art results based on language modeling. Recent works recorded eye movements of readers for the tasks of sentiment classification and sarcasm detection in sentiment text~\cite{mishra2016leveraging,mishra2018cognition}. The results show that words pertaining to a text's subjective summary attract a lot more attention. Mishra et al.~\cite{mishra2018cognition} proposed a novel multi-task learning approach for document-level sentiment analysis by considering the modality of human cognition along with text. 

\cm{In} contrast with movie reviews, Chen and Sokolova~\cite{chen2019unsupervised} investigated subjective (positive or negative) and objective terms in medical and scientific texts, showing that the majority of words for each dataset are objective (89.24\%), i.e., sentiment is not directly stated in the texts of these domains. 

Little work has been done on mining suggestions and tips until recently. \emph{SemEval-2019 Task 9} (see \cite{negi2019semeval}) was devoted to suggestion mining from user-generated texts, defining suggestions as sentences expressing advice, tips, and recommendations relating to a target entity. This task was formalized as sentence classification. Crowdsourced annotations were provided with sentences from software suggestion forums and hotel reviews from \emph{TripAdvisor}. Each text had \cm{been} annotated as a \textit{suggestion} or \textit{non-suggestion}. The authors presented two subtasks:
\begin{itemize}
\item \emph{Task A} for training and testing models on the dataset of software comments and
\item \emph{Task B} for testing models on hotel reviews without an in-domain training set.
\end{itemize}

As expected, the best systems used ensemble classifiers based on BERT and other neural networks. In particular, the winning team \emph{OleNet@Baidu}~\cite{liu2019olenet} achieved F-scores of $0.7812$ and $0.8579$ for subtasks A and B respectively. However, a rule-based classifier based on lexical patterns obtained an F-score of $0.858$ in subtask B~\cite{potamias2019ntua}. The \emph{SemEval-2019 Task 9} dataset shows that many suggestions from software forums include user requests, which is less frequent in other domains. 

Several studies in opinion mining have focused on the detection of complaints or technical problems in electronic products and mobile applications~\cite{iacob2013retrieving,khan2014no,ivanov2014clause,dong2013automated,tutubalina2015target}.
Khan et al.~\cite{khan2014no} described the ``No Fault Found'' phenomenon related to software usability problems. 
Iacob and Harrison~\cite{iacob2013retrieving} applied linguistic rules to classify feature requests, e.g., ``(the only thing) missing \textit{request}'', while Ivanov and Tutubalina~\cite{ivanov2014clause} proposed a clause-based approach to detect electronic product failures.
To sum up, detection of subjective and 
\cm{prescriptive}
texts (i.e., tips and suggestions) has not been extensively studied even though sentiment analysis is a very well researched problem with many different techniques.

\cm{As for natural language inference (NLI),} there exist three main sources of NLI datasets presented by Williams et al.~\cite{mnli}, Bowman et al.~\cite{bowman2015large}, and Khot et al.~\cite{khot2018scitail}. In these datasets, the basic NLI task is to classify the relation between two sentences (e.g., entailment). NLI has become popular in 
\nt{the last decade, especially rising in popularity with recently developed approaches that utilize}
Transformer-based models such as BERT~\cite{bert}, where it was adopted for few-shot and zero-shot learning along with text classification~\cite{yin2019benchmarking} and natural language understanding~\cite{kumar2017zero}. 
\add{The success of this NLI models-based line of work motivated our choice of the approach, which is why} 
we extend this methodology to suggestion detection.
\section{Data}\label{sec:data}
The dataset for suggestion mining used in this work was presented by Negi et al.~\cite{negi2019semeval} and used in \emph{SemEval2019, Task~9}\footnote{\url{https://github.com/Semeval2019Task9}}; we refer to~\cite{negi2019semeval} for a description of data collection and annotation \nt{procedures, which include a two-phase labeling procedure by crowd workers and experts respectively}.

In our setup we do not use the original validation (development) set but rather the so-called ``trial test set'', also released during the shared task run. Table~\ref{tab:dataset_stats} shows the statistics for this version of the dataset, \nt{including the training, validation, and testing parts of each dataset}. We also show samples from the dataset in Table~\ref{tab:dataset_samples}.

The data covers two independent domains.
\begin{enumerate}
    \item 
    \textit{Software suggestion forum}. Sentences for
this dataset were scraped from the Uservoice
platform that provides customer engagement tools to brands and hosts
dedicated suggestion forums for certain products.
Posts were scraped and split into
sentences using the Stanford CoreNLP toolkit.
Many suggestions are in the form of requests,
which is less frequent in other domains. The text
contains highly technical vocabulary related to the
software being discussed.
\item \textit{Hotel reviews}. Wachsmuth et al.~\cite{wachsmuth2014review} provide
a large dataset of \emph{TripAdvisor}
hotel reviews, split into statements so that each statement
has only one manually assigned sentiment label. 
Statements are equivalent to sentences, and consist of one or more clauses. \cite{negi2019semeval} annotated these segments as suggestions and non-suggestions.
\end{enumerate}

Following \cite{negi2019semeval}, we use two datasets, called Subtask~A and~B. In Subtask A the training, validation, and test parts are from the software forum domain, while Subtask B uses the same training part as Subtask A but with different test and validation parts. Since we consider the zero-shot setting, we do not use the training part.

\begin{table}[!t]
    \centering\footnotesize\setlength{\tabcolsep}{3pt}
    \begin{tabular}{|c|p{4.7cm}|p{2cm}r|p{2cm}r|}
     \hline
        \multicolumn{2}{|c|}{\textbf{Task/Domain}} & \multicolumn{2}{c|}{\textbf{Suggestions}} & \multicolumn{2}{c|}{\textbf{Non-Suggestions}} \\\hline
        \textbf{A} & Software development forums & 
        Training & 2085 & Training & 6415 \\
        & (Uservoice) & Validation & 296 & Validation & 296 \\
        & & Testing & 87 & Testing & 746 \\\hline
        \textbf{B} & Hotel reviews   & 
        Training & 2085 & Training & 6415 \\
        & (TripAdvisor) & Validation & 404 & Validation & 404 \\
        & & Testing & 348 & Testing & 476 \\\hline
    \end{tabular}
    \caption{Dataset statistics. The Train/validation/test split is shown for suggestions and non-suggestions for the two subtasks of~\cite{negi2019semeval}.}
    \label{tab:dataset_stats}

    \centering\setlength{\tabcolsep}{5pt}
    
    \begin{tabular}{|p{2.5cm}|p{5.0cm}|p{3.2cm}|}
    \hline
         & \textbf{Suggestion} & \textbf{Non-suggestion} \\\hline
        \textbf{SDE forums} & The proposal is to add something like:  // Something happened update your UI or run your business logic & I write a lot support ticket on this, but no one really cares on this issue. \\ \hline
        \textbf{Hotel reviews} & For a lovely breakfast, turn left out of the front entrance - on the next corner is a cafe with fresh baked breads and cooked meals. & A great choice!! \\\hline 
    \end{tabular}
    
    \caption{Samples from the SemEval2019-Task9 suggestion mining dataset: software development (SDE) forums and hotel reviews.}
    \label{tab:dataset_samples}
\end{table}

\section{Approach}\label{sec:methods}

Following the work on zero-shot learning for text classification benchmarking \cite{yin2019benchmarking}
\cm{and~using {\tt facebook/bart-large-mnli} (BART~\cite{lewis2019bart} trained on~\cite{mnli}) as a foundation model}, 
we have tried several different procedures for preparing labels in the \textit{label-fully-unseen} setting.

\paragraph{Approach 1.} In this approach we directly test whether the premise is suggesting something. The following statements are used as hypotheses: ``This text is a suggestion.'', ``This text is not a suggestion.'' We have also experimented with direct reformulations, e.g. ``This text suggests/is suggesting'' as well, with similar results
\nt{both on test and development sets}.

\paragraph{Approach 2.} For labels, we use the following definitions of ``suggestion'' from WordNet~\cite{miller1995wordnet,fellbaum1998wordnet}:
\begin{itemize}
    \item \emph{suggestion.n.01} (``This text is an idea that is suggested''),
    \item \emph{suggestion.n.02} (``This text is a proposal offered for acceptance or rejection''),
    \item \emph{suggestion.n.04} (``This text is persuasion formulated as a suggestion'').
\end{itemize}
\nt{Other definitions were discarded as irrelevant (\emph{trace.n.01}: a~just detectable amount, \emph{suggestion.n.05}: the~sequential mental process in~which one thought leads to~another by~association, \emph{hypnotism.n.01}: the~act of~inducing hypnosis).}
\cm{The non-suggestion label text used was as follows: ``This text is not a suggestion.''} 

Results of these simple approaches were arguably unsatisfactory. Each of the two classes (``suggestion'' and ``non-suggestion'') varies in the type of the possible message conveyed by the authors. Suggestion can be a plea, a question, a request etc., and non-suggestions are even more diverse: questions, comments, jokes, complaints and so on. 

\begin{table}[!t]
    \centering
    \begin{tabular}{|c|r|l|l|} \hline
        \textbf{?} & \textbf{Synset name} & \textbf{Lemma \# 1} & \textbf{Definition} \\ \hline
        
        & \textit{acknowledgment.n.03} & acknowledgment & a statement acknowledging something or someone \\ 
        & \textit{approval.n.04} & approval & a message expressing a favorable opinion \\ 
        & \textit{body.n.08} & body & the central message of a communication \\ 
        & \textit{commitment.n.04} & commitment & a message that makes a pledge \\ 
        & \textit{corker.n.01} & corker & (dated slang) a remarkable or excellent thing... \\ 
        & \textit{digression.n.01} & digression & a message that departs from the main subject \\ 
        \checkmark & \textit{direction.n.06} & direction & a message describing how something is... \\ 
        & \textit{disapproval.n.02} & disapproval & the expression of disapproval \\ 
        & \textit{disrespect.n.01} & disrespect & an expression of lack of respect \\ 
        & \textit{drivel.n.01} & drivel & a worthless message \\ 
        \checkmark & \textit{guidance.n.01} & guidance & something that provides direction or advice... \\ 
        & \textit{information.n.01} & information & a message received and understood \\ 
        & \textit{interpolation.n.01} & interpolation & a message (spoken or written) that is introduced... \\ 
        & \textit{latent\_content.n.01} & latent\_content & (psychoanalysis) hidden meaning of a fantasy... \\ 
        & \textit{meaning.n.01} & meaning & the message that is intended or expressed or... \\ 
        & \textit{narrative.n.01} & narrative & a message that tells the particulars of an act or... \\ 
        & \textit{nonsense.n.01} & nonsense & a message that seems to convey no meaning \\ 
        \checkmark & \textit{offer.n.02} & offer & something offered (as a proposal or bid) \\ 
        & \textit{opinion.n.02} & opinion & a message expressing a belief about something... \\ 
        \checkmark & \textit{promotion.n.01} & promotion & a message issued in behalf of some product or... \\ 
        \checkmark &  \textit{proposal.n.01} & proposal & something proposed (such as a plan or... \\ 
        & \textit{refusal.n.02} & refusal & a message refusing to accept something that... \\ 
        \checkmark & \textit{reminder.n.01} & reminder & a message that helps you remember something \\ 
        \checkmark & \textit{request.n.01} & request & a formal message requesting something that is... \\ 
        & \textit{respects.n.01} & respects & (often used with `pay') a formal expression of... \\ 
        & \textit{sensationalism.n.01} & sensationalism & subject matter that is calculated to excite and... \\ 
        & \textit{shocker.n.02} & shocker & a sensational message (in a film or play or novel) \\ 
        & \textit{statement.n.01} & statement & a message that is stated or declared; a communica... \\ 
        & \textit{statement.n.04} & statement & a nonverbal message \\ 
        & \textit{subject.n.01} & subject & the subject matter of a conversation... \\ 
        \checkmark & \textit{submission.n.01} & submission & something (manuscripts or architectural plans... \\ 
        & \textit{wit.n.01} & wit & a message whose ingenuity or verbal skill or... \\ \hline 
    \end{tabular}
    \caption{Hyponyms of WordNet synset \textit{message.n.02}. The possible candidates for direct mapping as a suggestion are checkmarked in column ``?''.}    \label{tab:messages}
\end{table}

\paragraph{Approach 3.} 
Given a wide variety of possible ``message types'', a natural idea would be to classify sentences into all of them and them map some labels to ``suggestions'' and others to ``non-suggestions''.

\nt{Thus, we have analyzed the place of ``suggestion'' in WordNet,
exploring the hyperonyms of ``suggestion''.}
The synset whose direct hyponyms in \emph{WordNet}~\cite{miller1995wordnet,fellbaum1998wordnet} described the types of the users' ``message'' and which we had found suitable for the task was the \emph{message.n.02} WordNet synset:
``what a~communication that is~about something is about''. \nt{A list of the direct hyponyms is presented in Table~\ref{tab:messages}}\footnote{See also \href{http://wordnetweb.princeton.edu/perl/webwn?o2=\&o0=1\&o8=1\&o1=1\&o7=\&o5=\&o9=\&o6=\&o3=\&o4=\&s=message\&i=2\&h=0100000\#c}{WordNet Web: wordnetweb.princeton.edu}.}.
Further, we have selected a list of candidates from those hyponyms to be later mapped to suggestions: 
\emph{direction.n.06}, \emph{guidance.n.01}, \emph{offer.n.02}, \emph{promotion.n.01}, \emph{proposal.n.01}, \emph{reminder.n.01}, \emph{request.n.01}, \emph{submission.n.01}.
We have formulated the labels as ``This~text is~a~[LEMMA]'', where [LEMMA] is the first lemma in the WordNet lemma synset list\footnote{We have also tried to enrich the labels by adding all lemmas joined by ``or'', but with no improvement; results of these experiments are given in \nt{Table~\ref{tab:extended_sets_dev_a}.}}.
We have used the development set of the \emph{SemEval2019 Task 9}, Subtask~A to find the best subset of candidate labels in terms of F1-measure of the ``suggestion'' class.

\section{Results}\label{sec:results}
\paragraph{Approaches 1 \& 2.} 
The results of ``asking'' entailment models whether the text is a suggestion or not are shown in Table~\ref{tab:simple_labels}. The difference in performance of all models on test sets
of Subtasks~A and~B
can be explained by the differences in target classes distributions; we also note
that Transformer-based models often struggle with negations~\cite{hossain-etal-2020-analysis}. 
Similar to~\cite{yin2019benchmarking}, using definitions seemingly does not guarantee better results for entailment-based zero-shot text (suggestion) classification.

\begin{table}[!t]
    \centering\footnotesize\setlength{\tabcolsep}{3pt}
    \begin{tabular}{|p{6cm}|c|cc|cc|}
        \hline 
        \multirow{2}{*}{\textbf{Premise}} & \multirow{2}{*}{\textbf{Task}}  &  \multicolumn{2}{c|}{\textbf{Dev. set}} & \multicolumn{2}{c|}{\textbf{Test set}} \\ \cline{3-6}
        & & F1 & Acc. & F1 & Acc. \\ \hline 
        ``This text is [not] a~suggestion.'' (A1) & A & 0.6727 & 0.5152 & 0.1961 & 0.1536 \\ 
         & B & 0.6616 & 0.5 & 0.5806 & 0.4163 \\ 
        \hline 
        ``This text is [not] suggesting.''(A1) & A& 0.6712 & 0.5118 &0.1898 & 0.1393 \\ 
         & B & 0.6617 & 0.4988 & 0.5840 &0.4175  \\
        \hline
        3 definitions VS &A & 0.6689 &0.5051 & 0.1925 & 0.1237 \\ 
        ``This text is \textbf{not} a~suggestion.'' (A2) &B & 0.6656 & 0.5025 & 0.5876 & 0.4175 \\
                      \hline 
        ``The best subset'' (A3) 
                    & A & 0.7517 &  0.7568 & 0.4479 & 0.8283 \\
                    & B & 0.4635 & 0.6361 & 0.4841 &  0.6699 \\ \hline
    \end{tabular}
    \caption{Results of Approaches 1 \& 2 and the best label subset from Approach 3 (Subtask~A).}
    \label{tab:simple_labels}\vspace{-.3cm}
\end{table}

\begin{figure}[!t]
    \centering
    \includegraphics[draft=false,width=0.65\linewidth]{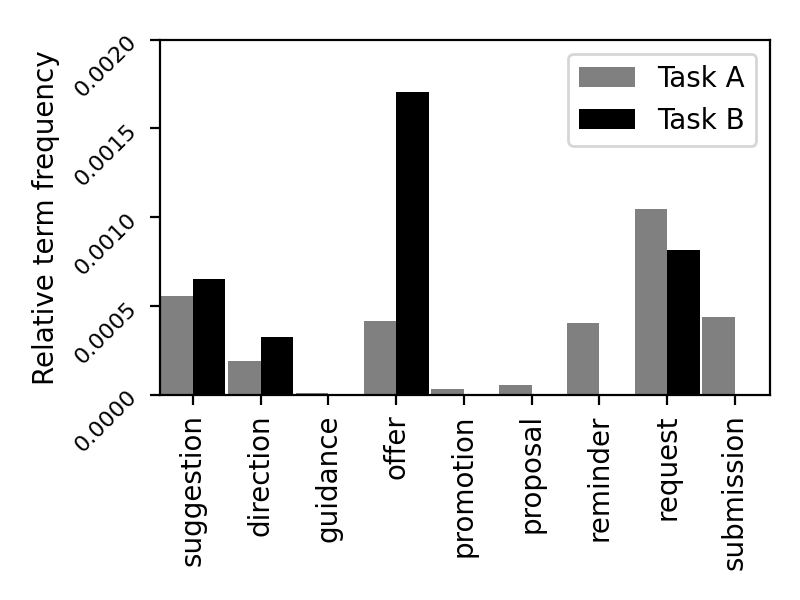}
    
    \caption{\cm{Relative word frequency in suggestions}
    in Subtask~A (software, train+dev+test) and Subtask~B (hotels, dev+test).}
    \label{fig:domain_differences}\vspace{-.3cm}
\end{figure}


\begin{table}[!t]\renewcommand{\arraystretch}{1.12}
    \centering\setlength{\tabcolsep}{2pt}
        \begin{tabular}{|c|p{.73\linewidth}|cc|}
        \hline
         \textbf{Size} &    \textbf{Labels subset} &    \textbf{F1} & \textbf{Acc.} \\  \hline\hline
         \multirow{3}{*}{4} & direction or instruction,proposal,reminder,request or petition or postulation & 0.6590 &   0.6976 \\\cline{2-4}
         &  direction or instruction,promotion or publicity or promotional\_material or packaging,proposal,request or petition or postulation & 0.6589 &   0.7027 \\\cline{2-4}
         &    direction or instruction,guidance or counsel or counseling or counselling or direction,promotion or publicity or promotional\_material or packaging,proposal & 0.6562 &   0.7027 \\ \hline\hline
            \multirow{3}{*}{5} &                                                                                                    direction or instruction,guidance or counsel or counseling or counselling or direction,proposal,reminder,request or petition or postulation & 0.6654 &   0.7010 \\\cline{2-4}
             &                                                 direction or instruction,guidance or counsel or counseling or counselling or direction,promotion or publicity or promotional\_material or packaging,proposal,request or petition or postulation & 0.6654 &   0.7061 \\\cline{2-4}
             &                                                                                                      direction or instruction,promotion or publicity or promotional\_material or packaging,proposal,reminder,request or petition or postulation & 0.6642 &   0.6993 \\ \hline\hline
            \multirow{3}{*}{6} &                                        direction or instruction,guidance or counsel or counseling or counselling or direction,promotion or publicity or promotional\_material or packaging,proposal,reminder,request or petition or postulation & 0.6704 &   0.7027 \\\cline{2-4}
             &                             direction or instruction,guidance or counsel or counseling or counselling or direction,promotion or publicity or promotional\_material or packaging,proposal,request or petition or postulation,submission or entry & 0.6679 &   0.7078 \\\cline{2-4}
             &                                                                                direction or instruction,guidance or counsel or counseling or counselling or direction,proposal,reminder,request or petition or postulation,submission or entry & 0.6679 &   0.7027 \\ \hline\hline
            \multirow{3}{*}{7} &                    direction or instruction,guidance or counsel or counseling or counselling or direction,promotion or publicity or promotional\_material or packaging,proposal,reminder,request or petition or postulation,submission or entry & 0.6729 &   0.7044 \\\cline{2-4}
             &                      direction or instruction,guidance or counsel or counseling or counselling or direction,offer or offering,promotion or publicity or promotional\_material or packaging,proposal,reminder,request or petition or postulation & 0.6704 &   0.6993 \\\cline{2-4}
             &           direction or instruction,guidance or counsel or counseling or counselling or direction,offer or offering,promotion or publicity or promotional\_material or packaging,proposal,request or petition or postulation,submission or entry & 0.6679 &   0.7044 \\ \hline\hline
            8 &  direction or instruction,guidance or counsel or counseling or counselling or direction,offer or offering,promotion or publicity or promotional\_material or packaging,proposal,reminder,request or petition or postulation,submission or entry & 0.6728 &   0.7010 \\
        \hline
        \end{tabular}

    \caption{Development set. Top-3 label (all lemmas concatenated with ``or'') \nt{extended} subsets results for each subset size from 5 to 8 selected by F1 measure.}
    \label{tab:extended_sets_dev_a}
\end{table}

\begin{table}[!t]\renewcommand{\arraystretch}{1.12}
    \centering
    \begin{tabular}{|c|r|cc|} \hline
 \textbf{Size} &                                                                   \textbf{Labels subset} & \textbf{F1} & \textbf{Accuracy} \\  \hline
    \multirow{3}{*}{4} &                                         guidance,offer,proposal,reminder & 0.6934 &   0.7351 \\
     &                                          offer,proposal,reminder,request & 0.6897 &   0.7327 \\
     &                                        direction,offer,proposal,reminder & 0.6802 &   0.7277 \\  \hline
    \multirow{3}{*}{5} &                                 guidance,offer,proposal,reminder,request & 0.7097 &   0.7438 \\
     &                               direction,guidance,offer,proposal,reminder & 0.7007 &   0.7389 \\
     &                                direction,offer,proposal,reminder,request & 0.6970 &   0.7364 \\  \hline
    \multirow{3}{*}{6} &                       direction,guidance,offer,proposal,reminder,request & 0.7167 &   0.7475 \\
     &                      guidance,offer,proposal,reminder,request,submission & 0.7123 &   0.7450 \\
     &                       guidance,offer,promotion,proposal,reminder,request & 0.7067 &   0.7401 \\  \hline
    \multirow{3}{*}{7} &            direction,guidance,offer,proposal,reminder,request,submission & \textbf{0.7192} &   0.7488 \\
     &             direction,guidance,offer,promotion,proposal,reminder,request & 0.7137 &   0.7438 \\
     &            guidance,offer,promotion,proposal,reminder,request,submission & 0.7093 &   0.7413 \\  \hline
    8 &  (all 8 labels) & 0.7163 &   0.7450 \\  \hline
\end{tabular}
    \label{tab:dev_b_top_3}
    \caption{Development set, Subtask B. Top-3 label subsets results for each subset size from $4$ to $8$ selected by F1 measure.}

    \centering
    \begin{tabular}{|c|r|cc|} \hline
 \textbf{Size} &                                                                   \textbf{Labels subset} & \textbf{F1} & \textbf{Accuracy} \\  \hline
    \multirow{3}{*}{4} &                                          offer,proposal,reminder,request & 0.5963 &   0.6845 \\
     &                                         guidance,offer,proposal,reminder & 0.5833 &   0.6723 \\
     &                                        offer,promotion,proposal,reminder & 0.5804 &   0.6772 \\ \hline
    \multirow{3}{*}{5} &                                 guidance,offer,proposal,reminder,request & 0.6030 &   0.6820 \\
     &                                offer,promotion,proposal,reminder,request & 0.6006 &   0.6869 \\
     &                               offer,proposal,reminder,request,submission & 0.5997 &   0.6857 \\ \hline
    \multirow{3}{*}{6} &                       guidance,offer,promotion,proposal,reminder,request & 0.6073 &   0.6845 \\
     &                      guidance,offer,proposal,reminder,request,submission & 0.6063 &   0.6833 \\
     &                       direction,guidance,offer,proposal,reminder,request & 0.6054 &   0.6820 \\ \hline
    \multirow{3}{*}{7} &            guidance,offer,promotion,proposal,reminder,request,submission & 0.6105 &   0.6857 \\
     &             direction,guidance,offer,promotion,proposal,reminder,request & 0.6096 &   0.6845 \\
     &            direction,guidance,offer,proposal,reminder,request,submission & 0.6087 &   0.6833 \\ \hline
    8 &  (all $8$ labels) & \textbf{0.6129} &   0.6857 \\
 \hline
\end{tabular}
    \label{tab:test_b_top_3}
    \caption{Test set, Subtask B. Top-3 label subsets results for each subset size from $4$ to $8$ selected by F1 measure.}
\end{table}

\paragraph{Approach 3.}
Results of the best subsets of labels (we evaluated all subsets of size $4$ to $8$) using the development set of Subtask~A are shown in Table~\ref{tab:dev_a_top_3}. The best subset is $\{\text{guidance}, \text{promotion}, \text{proposal}, \text{reminder}, \text{request}\}$, with F1-measure  $0.7517$ and accuracy $0.7568$. 
\nt{Half of the development set data points are annotated with ``suggestions'', while the test set is imbalanced (for details see Section~\ref{sec:data}), which may explain the difference in performance}.


\begin{table}[t]
    \centering\setlength{\tabcolsep}{3pt}\renewcommand{\arraystretch}{1.12}
    \begin{tabular}{|c|p{.7\textwidth}|cc|}
        \hline 
         \textbf{Size} &                                                                   \textbf{Subset of labels} &     \textbf{F1} & \textbf{Accuracy} \\  \hline\hline
             \multirow{3}{*}{4} &                                       guidance, proposal, reminder, request & 0.7509 &   0.7568 \\\cline{2-4}
             &                                      guidance, promotion, proposal, request & 0.7486 &   \textbf{0.7652} \\\cline{2-4}
             &                                          guidance, offer, proposal, request & 0.7455 &   0.7382 \\ \hline\hline
            \multirow{3}{*}{5} & 
            guidance, promotion, proposal, reminder, request& \textbf{0.7517} &   0.7568 \\\cline{2-4}
             &                                 guidance, offer, proposal, reminder, request & 0.7484 &   0.7297 \\\cline{2-4}
             &                                guidance, offer, promotion, proposal, request & 0.7463 &   0.7382 \\ \hline\hline
            \multirow{3}{*}{6} &                       guidance, offer, promotion, proposal, reminder, request & 0.7492 &   0.7297 \\\cline{2-4}
             &                   direction, guidance, promotion, proposal, reminder, request & 0.7462 &   0.7449 \\\cline{2-4}
             &                  guidance, promotion, proposal, reminder, request, submission & 0.7445 &   0.7449 \\  \hline\hline
            \multirow{3}{*}{7} &             direction, guidance, offer, promotion, proposal, reminder, request & 0.7443 &   0.7179 \\\cline{2-4}
             &            guidance, offer, promotion, proposal, reminder, request, submission & 0.7427 &   0.7179 \\\cline{2-4}
             &        direction, guidance, promotion, proposal, reminder, request, submission & 0.7393 &   0.7331 \\  \hline\hline
            8 &  All 8 labels  & 0.7380 &   0.7061 \\
        \hline 
        \end{tabular}
        
        \caption{Development set, \nt{Subtask A.} Top-3 label subsets results for each subset size from $4$ to $8$ selected by F1 measure.}
 \label{tab:dev_a_top_3}
\end{table}

We apply the same entailment-based zero-shot prediction procedure to the test sets of the two subtasks of \emph{SemEval2019, Task~9}. For the best combination of ``suggestion-related'' labels 
we achieve the following results.
\begin{itemize}
    \item[A.] F1-measure: $0.4479$, accuracy: $0.8283$. For~comparison, the best F1-measure in the supervised settings is $0.7812$~\cite{liu2019olenet}, while random uniform sampling yields mean F1-measure of $0.1734$.
    \item[B.] F1-measure: $0.4841$, accuracy: $0.6699$. Here, the best F1-measure in the supervised setting is $0.858$~\cite{potamias2019ntua}, and random sampling of labels yields $0.4566$. 
\end{itemize}
The results are also reported in Table~\ref{tab:simple_labels}. Clearly, F1-measures achieved for Subtask~B are not far from random prediction results for the label set tuned for Subtask~A.
We have also carried out the same entailment-based \nt{near-}zero-shot classification using all possible subsets of labels from size $4$ to size $8$ on Subtask~B development and test sets as well. \nt{ The $3$ best results for each label subset size are reported in Tables~\ref{tab:dev_b_top_3} and~\ref{tab:test_b_top_3}, respectively.}
The best results for Subtask~B are now much better: F1-measure on the development set is $0.7192$ (all labels except \emph{promotion})
and $0.6129$ on the test set (all $8$ labels).
Thus, the proposed method can achieve much better results with specially tuned label sets; this means that this approach is not entirely domain-independent. Figure~\ref{fig:domain_differences} supports this claim showing that the frequencies of words commonly used for suggestions are very different in hotel and software domains.

\section{Conclusion}\label{sec:conclusion}

In this work, we have evaluated several approaches to \textit{label-fully-unseen} zero-shot suggestion mining. We have proposed an approach based on hyponyms of the word \emph{message} from WordNet that outperforms direct labeling ``This text is [not] a suggestion'' and definitions of \emph{suggestion}
\nt{from \emph{WordNet}. 
However, choosing the best-performing subset of the hyponyms for one domain (software-related discussions) does not perform well on another (hotel reviews), which suggests that the subset of hyponyms is a new hyperparameter for the proposed approach. Either this hyperparameter needs to be tuned for each new domain, or new automated approaches for finding the best subset of hyponyms have to be developed, which we suggest as an interesting direction for further work.}

\FloatBarrier

\bibliography{naacl2021}
\bibliographystyle{splncs04}

\end{document}